\documentclass[10pt,conference]{IEEEtran}
\IEEEoverridecommandlockouts
\usepackage{cite}
\usepackage{float}
\usepackage{amsmath,amssymb,amsfonts}
\usepackage{graphicx, enumitem}
\usepackage{textcomp}
\usepackage[ruled,vlined,linesnumbered]{algorithm2e}
\usepackage[table,dvipsnames]{xcolor}
\usepackage{tikz}
\usepackage{bm}
\usepackage{color,soul}
\usepackage{makecell}
\usepackage{booktabs}
\usepackage{multirow}
\usepackage{url}
\usepackage{subfigure}
\usepackage{tikz}
\usepackage{listings}
\usepackage{xcolor}
\usepackage[none]{hyphenat}
\SetKwIF{If}{ElseIf}{Else}{if}{}{else if}{else}{end if}%
 
\setlist[itemize]{leftmargin=4mm}
\newcolumntype{L}[1]{>{\raggedright\let\newline\\\arraybackslash\hspace{0pt}}m{#1}}
\newcolumntype{C}[1]{>{\centering\let\newline\\\arraybackslash\hspace{0pt}}m{#1}}
\newcolumntype{R}[1]{>{\raggedleft\let\newline\\\arraybackslash\hspace{0pt}}m{#1}} 

\def\BibTeX{{\rm B\kern-.05em{\sc i\kern-.025em b}\kern-.08em
    T\kern-.1667em\lower.7ex\hbox{E}\kern-.125emX}}

\expandafter\def\expandafter\normalsize\expandafter{%
    \normalsize%
    \setlength\abovedisplayskip{4pt}%
    \setlength\belowdisplayskip{4pt}%
}

\begin{document}

\title{Quantum-Assisted Automatic Path-Planning for Robotic Quality Inspection in Industry 4.0
	{}
	\thanks{This work was supported by the Basque Government through ELKARTEK Program under Grants KK-2024/00024 (AURRERA project). The authors used Microsoft Copilot to enhance the language and clarity of the paper. The authors retain complete responsibility for the content of this research.}
}

\author{
	\IEEEauthorblockN{Eneko Osaba\IEEEauthorrefmark{2}\IEEEauthorrefmark{1},
        Estibaliz Garrote\IEEEauthorrefmark{2}\IEEEauthorrefmark{3},
        Pablo Miranda-Rodriguez\IEEEauthorrefmark{2}, \\
        Alessia Ciacco\IEEEauthorrefmark{4},
        Itziar Cabanes\IEEEauthorrefmark{3}, and
        Aitziber Mancisidor\IEEEauthorrefmark{3}
        }
	\IEEEauthorblockA{\IEEEauthorrefmark{2}TECNALIA, Basque Research and Technology Alliance (BRTA), 48160 Derio, Bizkaia, Spain}
    \IEEEauthorblockA{\IEEEauthorrefmark{3}Automatic Control and Systems Eng., Univ. of the Basque Country (UPV/EHU), Bilbao, Spain}
    \IEEEauthorblockA{\IEEEauthorrefmark{4}Department of Mechanical, Energy and Management Engineering, University of Calabria, Rende, Italy}
	\IEEEauthorblockA{\IEEEauthorrefmark{1}Corresponding author. Email: eneko.osaba@tecnalia.com}}
    
\maketitle

\IEEEpubidadjcol	

\begin{abstract}
This work explores the application of hybrid quantum-classical algorithms to optimize robotic inspection trajectories derived from  \textit{Computer-Aided Design} (CAD) models in industrial settings. By modeling the task as a 3D variant of the Traveling Salesman Problem—incorporating incomplete graphs and open-route constraints—this study evaluates the performance of two D-Wave-based solvers against classical methods such as GUROBI and Google OR-Tools. Results across five real-world cases demonstrate competitive solution quality with significantly reduced computation times, highlighting the potential of quantum approaches in automation under Industry 4.0.
\end{abstract}

\section{Introduction}\label{sec:intro}

Advances in quantum computing are enabling problem-solving capabilities at a scale beyond brute-force classical simulation~\cite{kim2023evidence}. As hardware improves—with more qubits, lower error rates, and faster execution—quantum algorithm research is advancing through both theory and experimentation. This progress is paving the way for increasingly sophisticated and compelling proof-of-concept demonstrations and applications, as well as the development of predominantly hybrid quantum-classical methods with growing potential \cite{abbas2024challenges}.

This study focuses on the quality inspection of industrial products. In particular, under the Industry 4.0 paradigm, the zero-defect requirement has driven the need to implement adaptable systems capable of inspecting evolving manufactured parts. The integration of machine vision and robotics for quality control provides the flexibility required to adapt to changes in production, such as new, updated, or customized products. To enhance the level of automation, a new method for the automatic generation of robot trajectories from the product's \textit{Computer-Aided Design} (CAD) file is proposed in \cite{garrote2024automated}. In this method, the final trajectory is determined using an optimization algorithm.

The present work addresses this optimization task with quantum methods. Five references has been selected for validation, each corresponding to a different production process: a car door, a car rear bumper, a toy bear, a model airplane, and a half-sphere. We employ two solvers from D-Wave's \textit{Hybrid Solver Service} \cite{HSS}—the \textit{Leap Constrained Quadratic Model Hybrid Solver} (\texttt{CQM-Hybrid}) and the \textit{Nonlinear-Program Hybrid Solver} (\texttt{NL-Hybrid})—emphasizing the latter due to its superior performance. Results are benchmarked against two classical methods: GUROBI and Google OR-Tools.

\section{Methods}\label{sec:method}

The problem outlined above has been modeled as a trajectory optimization task, based on a modified version of the Traveling Salesman Problem (TSP). We have incorporated three key features: the three-dimensional nature of the points, the use of an incomplete graph, and the consideration of open routes (i.e., the route starts and ends at different locations). %It is worth noting that, while similar problems have been explored in the literature from the classical perspective \cite{ganganath2016trajectory}, they have not yet been approached from a quantum perspective, making this a significant contribution of our work.

As mentioned earlier, although we explored two algorithmic approaches—\texttt{CQM-Hybrid} and \texttt{NL-Hybrid}, extensive testing revealed that \texttt{NL-Hybrid} consistently delivered superior results. Therefore, the implementation described here refers to this method. For those interested in the development details, all source code is publicly available\footnote{\url{http://dx.doi.org/10.17632/n7mrwz3fgn.2}}.

The \texttt{NL-Hybrid} is a hybrid quantum-classical algorithm that naturally supports nonlinear constraints—whether linear, quadratic, or higher-order—expressed as equalities or inequalities. In essence, it operates in two phases: first, it initializes a set of parallel hybrid threads, each combining a \textit{Classical Heuristic Module} to explore the problem's solution space, and a \textit{Quantum Module} that sends quantum queries to D-Wave’s \texttt{Advantage\_system6.4} QPU. These queries guide the heuristic module toward more promising regions of the search space or help refine existing solutions. For additional details on \texttt{NL-Hybrid}, we refer to interested readers to our previous work \cite{osaba2025d}.

The TSP variant addressed in this work is well-suited for the \texttt{NL-Hybrid} solver, as its decision variables can be directly defined using the native \texttt{list(number\_variables)} type, representing a permutation of the problem-nodes. This permutation-based encoding eliminates the need for additional constraints, unlike formulations such as QUBO.

The following Python snippet demonstrates how to initialize the model and decision variables, where $N$ is the number of nodes and $C$ is an $N \times N$ cost matrix with $c_{ij}$ values (possibly incomplete, as the graph is not fully connected). It also shows how to define the problem and set the objective function, mathematically expressed as $f(\mathbf{x}) = \sum_{i=1}^{N-1}cost\_matrix_{x_i,x_{i+1}}$, with $\mathbf{x}$ being a feasible \texttt{path} represented as a \texttt{list}.

\begin{lstlisting}
from dwave.optimization.model import Model
problem_model = Model()
# Defining the variable as a list of size N
path = problem_model.list(N)
# Entering the cost matrix as constant 
cost_matrix = problem_model.constant(C)
# Sum the costs of the full path
path_cost = cost_matrix[path[:-1], path[1:]]    
# The objective is introduced using model.minimize
problem_model.minimize(path_cost)
\end{lstlisting}
Once the problem is modeled as shown in this code snippet, it is ready to be solved using \texttt{NL-Hybrid}.

\section{Experimentation}\label{sec:exp}

The experimentation has been conducted on five different real-world instances. The data for the cases have been calculated as part of the automatic trajectory generation process for the complete surface inspection by a machine vision system mounted on a robot. The inspection points were determined by segmenting each reference surface into patches, applying two constraints: the area of each patch and the variation in the surface normal within the patch must both be below predefined thresholds. The cost between pairs of points is defined by a function that measures the robot's cost to move from one point to another. 

The objective of the tests is to evaluate the performance of \texttt{NL-Hybrid} and \texttt{CQM-Hybrid} in terms of solution quality and runtime. Furthermore, to properly contextualize these results, we compared them with two widely used classical methods. The first is Google OR-Tools, for which a variation of the default configuration was developed to address the specific problem in this study\footnote{https://developers.google.com/optimization/routing/tsp}. The second is GUROBI\footnote{https://www.gurobi.com/solutions/gurobi-optimizer/}, a powerful exact solver frequently used as a baseline in quantum computing. In fact, GUROBI delivers the best results in terms of solution quality; therefore, we use its outcomes as a reference for the evaluation metric employed: the approximation ratio (\textit{AR}), defined as the ratio between the solution obtained by a solver and the baseline solution.

The results obtained by all the methods are presented in Table \ref{tab:results}, based on 15 independent runs. The table reports the average runtime (\textit{rt}), in seconds, and the average quality of the best solution found in each run, considering that each method may produce a finite set of outcomes per execution. The table also includes the size characteristics of each benchmark instance, specifically the number of nodes and edges.

\begin{figure}[t]
 \centering
 \includegraphics[width=0.90\linewidth]{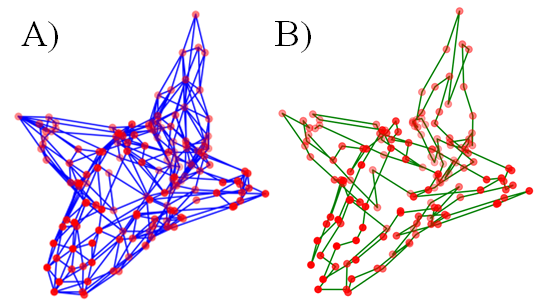}
 \caption{Aircraft Instance. (a) the graph representing the instance, including all nodes and existing edges; (b) a possible solution offered by the \texttt{NL-Hybrid}.}
 \label{fig:aircraft}
\end{figure}

The outcomes reveal several insights. Among hybrid algorithms, \texttt{NL-Hybrid} clearly outperforms \texttt{CQM-Hybrid} in both \textit{AR} and \textit{rt}. This trend also holds when compared to OR-Tools, which is particularly noteworthy given the recognition this solver has received in the related literature~\cite{benoit2024navigating}. Finally, while \texttt{NL-Hybrid} does not surpass GUROBI in solution quality, it offers competitive performance with significantly reduced computation time—which can be critical in industrial applications. To support the visualization of the results presented in this study, Figure \ref{fig:aircraft} shows the \textit{aircraft} instance along with a possible solution generated by \texttt{NL-Hybrid}. For interested readers, the full set of results and all graphical representations are publicly accessible$^1$.

In conclusion, this research represents a further example of the current potential of quantum computing in addressing industrial problems. To reinforce the findings presented, additional experiments are planned for the near future.

\begin{table}[t]
  \caption{AR and rt values obtained by all methods across the benchmark instances. NL-H = NL-Hybrid. CQM-H = CQM-Hybrid. OR-T = OR-Tools. GUR = GUROBI.}
  \label{tab:results}
    \centering
    \resizebox{1.0\columnwidth}{!}{
        \begin{tabular}{c|ll|ll|ll|l}
            \toprule
            \multirow{2}{*}{\textbf{Instance} (nodes, edges)} & \multicolumn{2}{c|}{NL-H} & \multicolumn{2}{c|}{CQM-H} & \multicolumn{2}{c|}{OR-T} & \multicolumn{1}{c}{GUR}\\
            & AR & rt & AR & rt & AR & rt & rt \\
            \midrule
            \textbf{Car-Door} (106, 184) & 0.93 & 4.9 & 0.75 & 1.2K & 0.83 & 6.8 & 17.4\\
            \textbf{Bumper} (116, 253) & 0.94 & 5.1 & 0.73 & 1.4K & 0.79 & 7.2 & 804.5\\
            \textbf{Toy-Bear} (149, 397) & 0.88 & 5.2 & 0.65 & 3.2K & 0.75 & 7.8 & 2.4K\\
            \textbf{Aircraft} (151, 441) & 0.86 & 4.5 & 0.67 & 3.6K & 0.69 & 9.4 & 11.8K\\
            \textbf{H-Sphere} (194, 393) & 0.83 & 5.4 & 0.62 & 6.8K & 0.79 & 13.5 & 68.4K\\
            \bottomrule
        \end{tabular}
    }
\end{table}

\bibliographystyle{IEEEtran}
\bibliography{IEEEexample}
\end{document}